\documentclass{article}

\usepackage{arxiv}

\usepackage[utf8]{inputenc} 
\usepackage[T1]{fontenc}    
\usepackage{hyperref}       
\usepackage{url}            
\usepackage{booktabs}       
\usepackage{amsfonts}       
\usepackage{nicefrac}       
\usepackage{microtype}      
\usepackage{lipsum}
\usepackage{graphicx}
\usepackage{footnote}
\usepackage[flushleft]{threeparttable}
\usepackage{amsmath}
\usepackage{color}
\usepackage{float}
\usepackage{subcaption}
\usepackage{flushend}
\usepackage{wrapfig}
\newsavebox{\measurebox}
\graphicspath{ {./images/} }

\title{\LARGE \bf Characterization of Assistive Robot Arm Teleoperation: \protect\\A Preliminary Study to Inform
Shared Control}

\author{
 Mahdieh Nejati Javaremi \\
  Northwestern University\\
  Shirley Ryan AbilityLab \\
  Chicago, IL, USA \\
  \texttt{m.nejati@u.northwestern.edu} \\
   \And
 Brenna D. Argall \\
  Northwestern University\\
  Shirley Ryan AbilityLab\\
  Chicago, IL, USA \\
  \texttt{brenna.argall@northwestern.edu} \\
}

\begin{document}
\maketitle
\begin{abstract}
Assistive robotic devices can increase the independence of individuals with motor impairments. However, each person is unique in their level of injury, preferences, and skills, which moreover can change over time. Further, the amount of assistance required can vary throughout the day due to pain or fatigue, or over longer periods due to rehabilitation, debilitating conditions, or aging. Therefore, in order to become an effective team member, the assistive machine should be able to learn from and adapt to the human user. To do so, we need to be able to characterize the user's control commands to determine when and how autonomy should change to best assist the user. We perform a 20 person pilot study in order to establish a set of meaningful performance measures which can be used to characterize the user's control signals and as cues for the autonomy to modify the level and amount of assistance. Our study includes 8 spinal cord injured and 12 uninjured individuals. The results unveil a set of objective, runtime-computable metrics that are correlated with user-perceived task difficulty, and thus could be used by an autonomy system when deciding whether assistance is required. The results further show that metrics which evaluate the user interaction with the robotic device, robot execution, and the perceived task difficulty show differences among spinal cord injured and uninjured groups, and are affected by the type of control interface used. The results will be used to develop an adaptable, user-centered, and individually customized shared-control algorithms.  
\end{abstract}

\section{Introduction}

Assistive devices such as robotic arms and powered wheelchairs are designed to return independence to people with motor impairments such as spinal cord injury (SCI), amyotrophic lateral sclerosis (ALS), and cerebral palsy (CP), among others. Controlling assistive devices can be challenging---especially when trying to control more complex assistive devices with more limited interfaces~\cite{Cowan2012}.

Commercially-available interfaces used for controlling assistive devices can be either proportional---where the user has control over both \textit{which} control signal and its \textit{magnitude}---or discrete---where the user only has control over the selection of \textit{which} signal to turn on or off. Common control interfaces can be one, two, or three dimensional. To fully control all six Degrees-of-Freedom (DoF) of a robotic arm's end-effector, which includes the three linear and three angular positions in space, the user must switch between subsets of the control dimensions. The subsets are referred to as \textit{control modes}. With a one-dimensional interface, the user must switch between six modes to fully control all 6-DoF. This increases the physical and cognitive burden on the user. 

Autonomy can help offset the challenges that exist in operating these interfaces~\cite{fehr2000}. A previous study with SCI users of powered wheelchairs found that they prefer to retain as much control over the assistive machine as possible~\cite{gopinath2017human}. Sharing control between the user and autonomy can allow the user to retain control authority while benefiting from the assistance of autonomy when the task becomes too burdensome~\cite{argall2013}. However, there is no one-size-fits-all method of control sharing as each person is unique in their desired control preference~\cite{erdogan:2017:ras}. Their motor abilities and required level and type of assistance also can change. For control sharing to be useful, practical and accepted in the realm of assistive robots for motor-impaired users, it is furthermore important for the control-sharing to be able to adapt to new scenarios and user skills~\cite{Urdiales2009}. 

In this study, we investigate different performance measures to quantify spinal cord injured and uninjured user teleoperation characteristics. Specifically, we investigate these different performance measures for their potential as useful cues for when autonomy should be triggered to be most helpful to the user. These measures are important as they can furthermore inform the design of control sharing paradigms that are able to adapt to the user's variability over time. We also investigate differences between different subject groups and determine whether subject specific training data is needed for algorithm development. Towards this, we conduct a 20 person pilot study with 8 SCI and 12 uninjured subjects using a robotic arm to accomplish various Activities of Daily Living (ADL) tasks using three different types of commercially-available interfaces. We determine relationships between the selection of performance measures and assess the strength of correlations between these and the user's perceived task difficulty. Our premise is that the information content encoded within the user's control signals has useful characteristics and implications for algorithms capable of learning from and adapting to the user.  

In Section~\ref{sec:background} we cover related work on control sharing in assistive robotics. In Section~\ref{sec:exp_design} we provide a detailed description of our experimental design. We present our results in Section~\ref{sec:results} followed by a discussion of our findings in Section~\ref{sec:discussion}. Finally we 
present our conclusions and future work in Section~\ref{sec:conclusion}. 
\section{Background}
\label{sec:background}
In this section we overview related work in existing methods of shared control and user signal characterization in the domain of assistive robotics. 

There are various examples of different control sharing algorithms in the literature~\cite{music:2017:arc}, which can largely be classified based on the way control is allocated between the user and autonomy. In general, control-sharing algorithms can be divided into those where (a) the amount or \textit{level} of control authority is shared between the human and the robot, and (b) those where there is an idea of \textit{subtask} allocation. In addition to various ways of sharing control between humans and robots in assistive settings, there are also many ways in which shared control can be triggered. 

In schemes where the level of autonomy is shared, the amount of control shared by autonomy can be static or dynamic~\cite{beer:2014:hri}. The dynamic change in the control authority is typically either implicit based on a change in the environment~\cite{Simpson1998}, or explicitly set by the human user~\cite{parikh:2004}. 

In schemes where there is an idea of subtask allocation, there are specific behaviors where the robot will take control authority---including but not limited to safe stopping at obstacles or obstacle avoidance and manipulation of grasped objects. These subroutines can either be always on, or triggered explicitly by the end-user~\cite{bourhis:2001}, or implicitly by a change in the environment or the robot state~\cite{lankenau:2001}. 

In order for control-sharing to be useful and effective for motor-impaired end-users, these methods must be able to adapt to the end-user---whether it be in the appropriate level of relinquishing control to autonomy at a given time, or when to trigger autonomy to take over certain subtasks. Some research has focused on adapting the control sharing to changes in a powered wheelchair operator's environment and skilled user intent~\cite{Soh2015}. Other research also looked at adaptable learning from expert assistants (i.e. occupational therapists) and not the actual end-user~\cite{Kucukyilmaz2015}. 

Limited work has been done that considers variation in user skill. Vanhooydonck \textit{et al.} use a neural network to model a personalized shared-assistance wheelchair system~\cite{Vanhooydonck2010}. However, the model relies on gathering a large number of training data for any new condition---including changes in skill level or the environment---and cannot adapt to changes online.  Fdez-Carmona \textit{et al.} propose a navigation skill profile to trigger the adaptive assistance of a powered wheelchair~\cite{Fdez-Carmona2014}. They use smoothness and directness to define the skill profile. In addition to smoothness and directness, safety and disagreement have also been used as factors to characterize user signal while operating a powered wheelchair~\cite{Urdiales2013}. To our knowledge, there has been no previous work done on characterizing teleoperation control signals of people operating an assistive robotic arm. 

Smoothness and directness are characteristics of the user command signal. In the domain of shared-control assistive robots, both the user and the robot signals can be characterized. There are also examples of user signal characterization outside the domain of assistive robotics. In sensory-motor performance research, movement and trajectory smoothness is often used to measure performance skill~\cite{Hogan2009}. In the domain of automotive engineering, parameters such as steering angle and frequency of user inputs are used to characterize driving behavior~\cite{Zuojin2017, Fernandez2016}.  

We aim in this work to identify computable measures to sufficiently represent characteristics of the user's operation of an assistive robot, with the long term goal of using these measures within algorithms that can adapt to the user's unique skill level, needs, and changing control behavior. Towards this end, we identify performance measures which can be useful as real-time predictors of changes in perceived user difficulty. \par

\section{METHODS}\label{sec:exp_design}
This section provides a detailed description of the research design and procedures used in this experiment.  
\subsection{Hardware}
The study was conducted using the MICO robotic arm (Kinova Robotics, Canada). This platform was chosen because it is a commercially available assistive device designed for users of powered-wheelchairs. 

\begin{figure}[t]
\centering
\vspace{0.2cm}
\includegraphics[height=2.2cm]{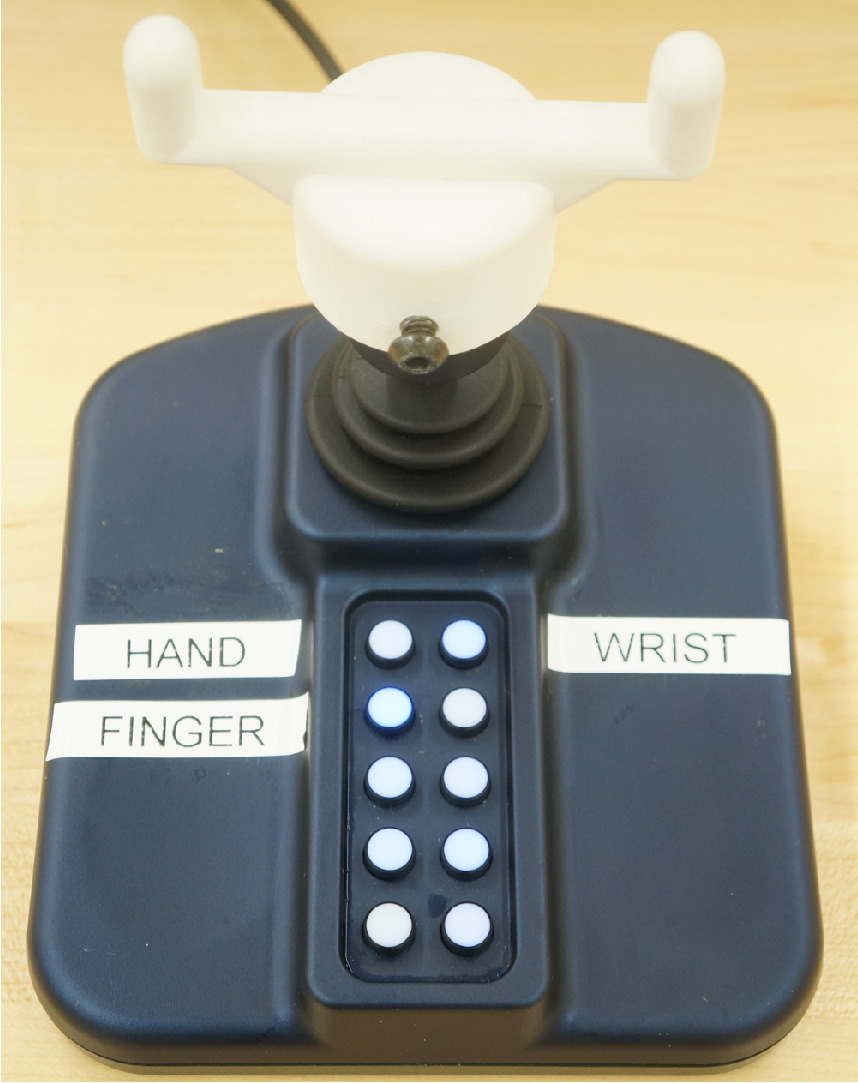}
\includegraphics[height=2.2cm]{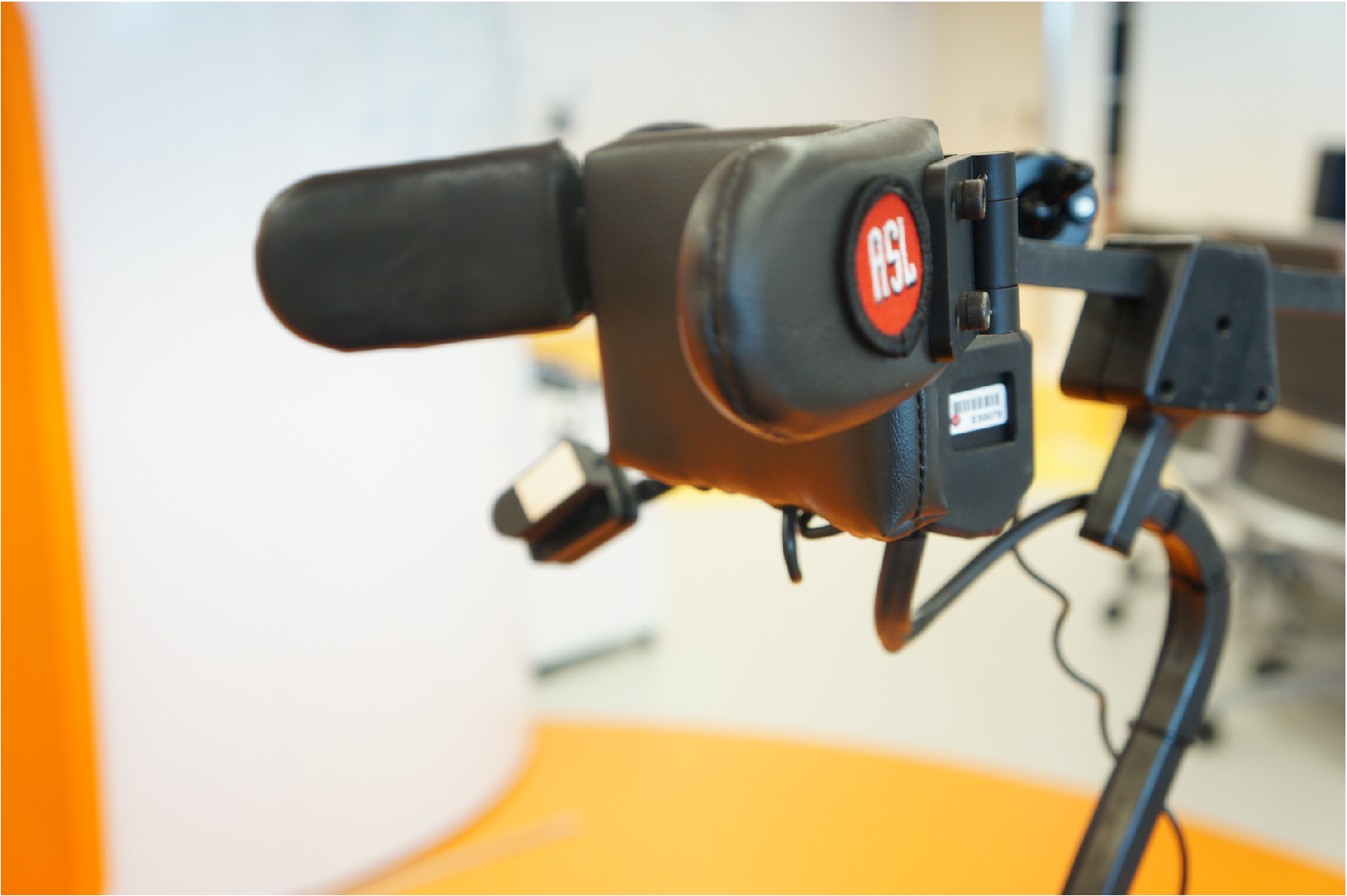}
\includegraphics[height=2.2cm]{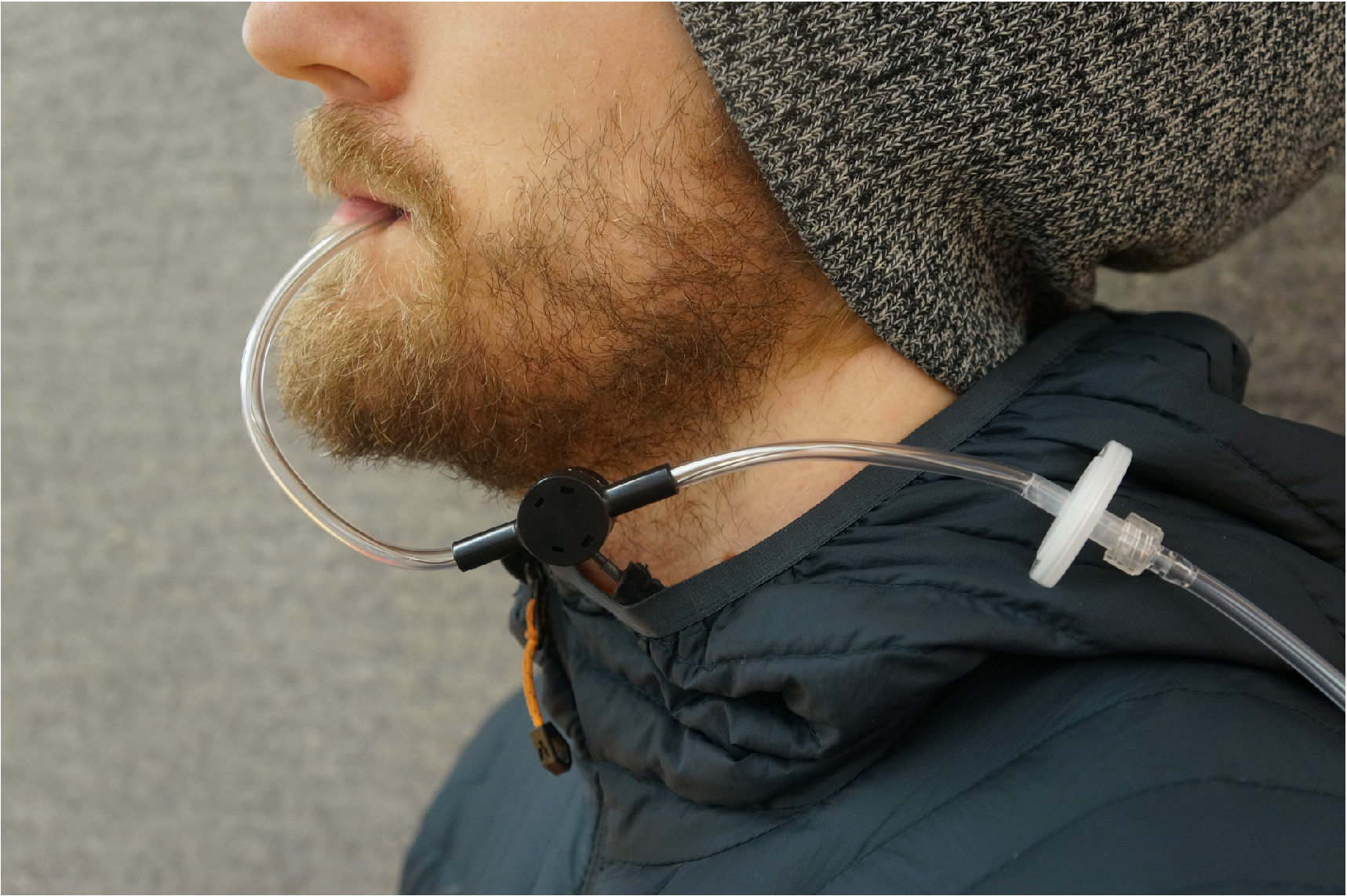}
\caption{Control interfaces used in this study. \textit{From left to right:} 3-axis joystick, head array system, sip/puff switch.}
\label{fig:interfaces}
\end{figure}

Each subject was asked to perform two different tasks using three different interfaces (Fig.~\ref{fig:interfaces}): (1) IPD Ultima 3-axis joystick (CH-Products Industrial, CA, USA), which is a three dimensional proportional controller, (2) Sip/Puff switch (Origin Instruments, TX, USA), which is a one dimensional proportional controller, and (3) 105 electronic head array system (ASL, TX, USA), which is a one dimensional discrete controller. These interfaces were selected as they are commonly used interfaces employed by SCI users of powered wheelchairs. 

Four wearable BioStampRC sensors (MC10 Inc., MA, USA) were used to measure physiological signals. One sensor was configured as an ECG sensor to measure heart rate. The other three were configured as accelerometers to measure hand, head, and torso kinematics. This selection was made because these sensors are wireless and therefore do not limit the human-user's motion.  

\subsection{Participants}
The study consisted of 20 participants: 12 uninjured and 8 with spinal cord injury (levels C3, C5-C6, C5-C7 incomplete, C6 incomplete, C5-C6, C3 incomplete, C3-C4, T2-3). From the spinal cord injured group, one participant was not able to use the joystick due to no upper-limb control, and one participant required a neck brace and was not able to use the head array switch. The age and gender demographics are provided in Table~\ref{table:demog}.

\begin{figure*}[h]
\centering
\vspace{0.2cm}
\begin{subfigure}[b]{.32\linewidth}
\includegraphics[width=\linewidth]{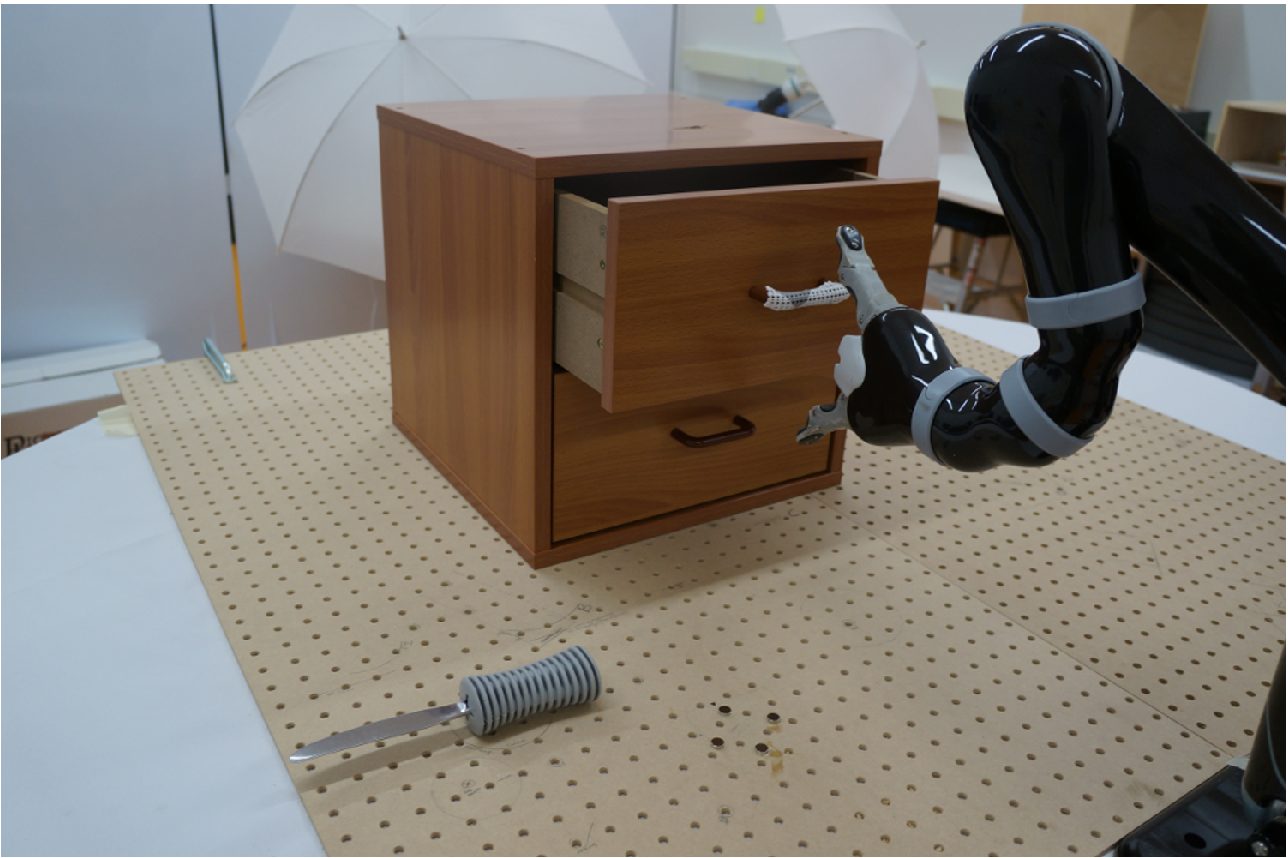}
\caption{}\label{fig:task1}
\end{subfigure}
\begin{subfigure}[b]{.32\linewidth}
\includegraphics[width=\linewidth]{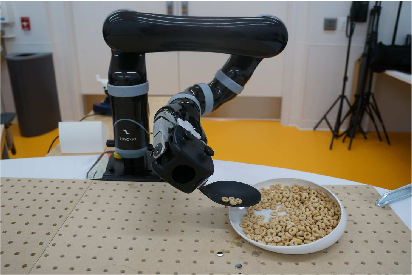}
\caption{}\label{fig:task2}
\end{subfigure}
\begin{subfigure}[b]{.32\linewidth}
\includegraphics[width=\linewidth]{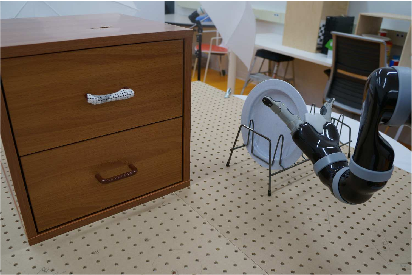}
\caption{}\label{fig:task3}
\end{subfigure}

\begin{subfigure}[b]{.32\linewidth}
\includegraphics[width=\linewidth]{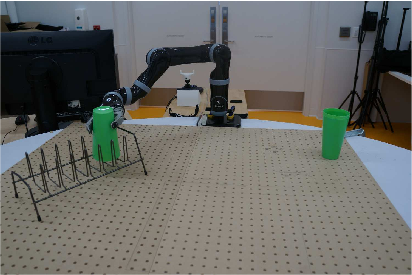}
\caption{}\label{fig:task4}
\end{subfigure}
\begin{subfigure}[b]{.32\linewidth}
\includegraphics[width=\linewidth]{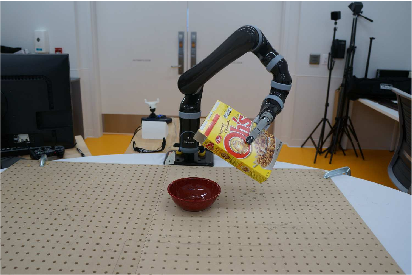}
\caption{}\label{fig:task5}
\end{subfigure}
\begin{subfigure}[b]{.32\linewidth}
\includegraphics[width=\linewidth]{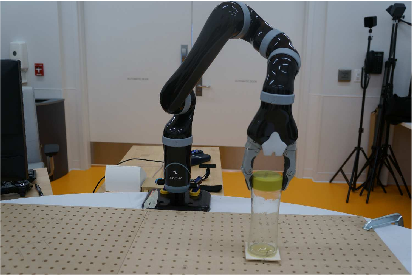}
\caption{}\label{fig:task6}
\end{subfigure}

 \caption{The six ADL tasks used in the experiment:
    (\subref{fig:task1}) Open drawer, pick up butter knife, place it inside the drawer.
    (\subref{fig:task2}) Scoop cereal without spilling, position spoon towards mouth.
    (\subref{fig:task3}) Pick up plate from top of drawers, place in dish rack.
    (\subref{fig:task4}) Pick up inverted cup from the dish rack, rotate and stack inside upright cups. 
	  (\subref{fig:task5}) Grasp cereal box, pour into bowl without spilling.
	  (\subref{fig:task6}) Unscrew jar lid, lift lid straight up.}
\label{fig:tasks}
\end{figure*}

\begin{table}[H]
\centering
\caption{Demographics of Study Participants.}
\label{table:demog}
\begin{tabular}{ |c|c|c| } 
\cline{2-3}
 \multicolumn{1}{c}{} & \multicolumn{1}{|c|}{SCI} & Uninjured \\
\hline
Number & 8 & 12 \\ \hline
Age (mean$\pm$std) & 45.3$\pm$16.7 & 26.9$\pm$3.64 \\ \hline
Female-Male& 1-7 & 6-6 \\ 
\hline
\end{tabular}
\end{table}

\subsection{Procedure}
The protocol and consent form were approved by Northwestern University's Internal Review Board (IRB) and all participants gave informed signed consent. 

\subsubsection{Protocol}
Each session began by obtaining the participant's consent, demographics information, and completing a pre-session questionnaire. Then the BioStampRC sensors were attached and the participant's baseline signals were measured. A session consisted of three rounds---one per interface. Prior to starting the first trial in each round, the participants were given time to train with each new interface. The subjects then performed two separate tasks twice. Immediately after the second execution of each task, the participants completed post-task questionnaires to self-report task difficulty. The order of the tasks and interfaces were randomly counterbalanced across all subjects prior to recruitment, and each participant was randomly assigned conditions. The session ended by measuring the resting heart rate signal and a final questionnaire comparing interfaces and tasks. We collected a total of 222 trials: 78 spinal cord-injured (26 per interface) and 144 uninjured (48 per interface). 

\subsubsection{Tasks}
A selection of six different ADL tasks were designed with the theme of eating and housekeeping (Fig~\ref{fig:tasks}). Tasks were selected in consultation with an occupational therapist. Each participant was assigned two tasks from this selection in a counterbalanced manner. Each subject performed their two tasks with all three interfaces. There were 15 unique sets of task conditions, and no two subjects performed the same combination of tasks.  

\subsection{Experimental Design}
The experiment used a 2x3 mixed design, where the interface was considered a \textit{within-subjects} factor and whether or not the participant was uninjured or had a spinal-cord injury was a \textit{between-groups} factor. In this preliminary study we investigated task-independent characteristics of user control commands. Therefore, we did not include the selection of task as a between-subjects factor in the experimental design.

\subsection{Metrics}
The metrics detailed in the following section were the dependent variables analyzed for their significance in characterizing the user signals. 
\subsubsection{Subjective}
\begin{itemize}
    \item Task difficulty: The user's perception of how difficult it was to complete the task using the robotic arm and control interface. This was measured using a post-task Likert scale questionnaire, which was completed immediately after task execution and included the following three statements: 
    \begin{itemize}
    \item Task Difficulty: It was easy for me to complete this task. 
    \item Robot Difficulty: The robot was easy for me to operate. 
    \item Interface Difficulty: I was able to issue my intended control command. 
    \end{itemize}
    The overall difficulty is the sum of the three Likert scores. 
\end{itemize}

\subsubsection{Physiological}
\begin{itemize}
    \item Heart rate variability: The time-difference between successive heart beats. The heart rate was measured at a frequency of 250 Hz. The time-domain Root Mean Square of the Successive Differences (RMSSD)~\cite{Munoz2015} was used to assess the heart rate variability:  
\begin{equation*}
    RMSSD=\sqrt{\frac{1}{N-2}\sum_{n=3}^{N}[I(n)-I(n-1)]^2}
\end{equation*}
where $N$ is the total number of heart beats and $I(\cdot)$ is the interbeat interval.
\end{itemize}

\subsubsection{User interaction}
\begin{itemize}
    \item Frequency of input commands: The rate at which the user issued a command through the interface. We used the Exponential Moving Average (EMA) $f_t$ over a window size of 10 input commands, 
    \begin{align*}
    Y_t &= \frac{1}{10}\sum_{i=N_{t-10}}^{N_t}\frac{1}{t_i-t_{i-1}} \\
    f_t &= \alpha \cdot Y_t + (1-\alpha) \cdot f_{t-1}
  \end{align*}
  where $N_t$ is the total number of commands in the window, $Y_t$ is the input rate at time $t$, $f_{t-1}$ is the value of the EMA at time $t-1$, and $\alpha$ is the degree of weight decrease. 
  
    \item Number of mode switches: The number of times the user switched between control modes.
    
    \item Smoothness of user input commands: The quality of continuity or non-intermittency of input commands.

We measured smoothness using the Spectral Arc Length (SPARC)~\cite{Balasubramanian2012}: \\ 
  \begin{align*}
          SPARC &= - \int_{0}^{\omega_c}\bigg[\Big(\frac{1}{\omega_c}\Big)^2+\Big(\frac{d\hat{V}(\omega)}{d\omega}\Big)^2\bigg]^\frac{1}{2}d\omega; \\
          \hat{V}(\omega)&=\frac{V(\omega)}{V(0)}\\
    \omega_c &= min(\omega_c^{max}, min(\omega, \hat{V}(r) < \bar{V} ~\forall ~r>\omega))
  \end{align*}
  where $V(\omega)$ is the Fourier magnitude spectrum of user input velocity $\omega$, $\hat{V}(\omega)$ is the normalized spectrum of user input velocities, and $\omega_c$ is the dynamic cutoff frequency which determines sensitivity to noise. This metric was chosen because it was demonstrated to be a reliable measure of smoothness that is not affected by the amplitude or duration of input signals~\cite{Balasubramanian2015}. 
\end{itemize}

\subsubsection{Robot execution} 
\begin{itemize}
\item Trajectory smoothness: The continuity of the end-effector's trajectory. We measure the smoothness of the end-effector trajectories using SPARC, where $V(\omega)$ is the Fourier magnitude spectrum of end-effector velocities, and $\hat{V}(\omega)$ is the normalized spectrum of end-effector velocities.
\end{itemize}

\section{RESULTS}\label{sec:results}
For each metric, the Shapiro-Wilk test of normality was performed to ensure all dependent variables were approximately normally distributed. Both group sizes and variances were unequal, therefore we used a general linear model (GLM) to determine any inter-group significant interactions. For within-group interactions, we used Tukey's multiple comparison procedure. We used the Pearson's correlation coefficient to determine the strength and direction of any correlations between task difficulty and the objective metrics (R-value in Tables~\ref{table:corr} and~\ref{table:perf}). 

\clearpage 
\subsection{Subjective Measure}\label{sec:results_subjective}
\begin{wrapfigure}{r}{0.35\textwidth}
\begin{center}
\includegraphics[width=0.3\textwidth]{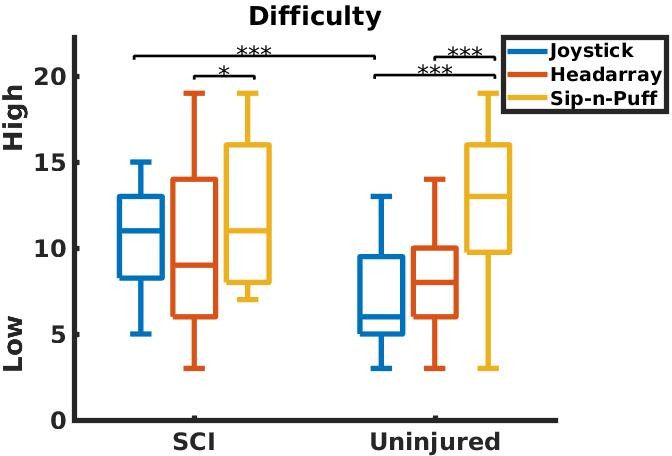}
\caption{Analysis of user-perceived difficulty between all subjects and interfaces.}
\label{fig:difficulty}
\end{center}
\end{wrapfigure}

Our preliminary results show that the user's perceived difficulty in the uninjured group is strongly dependent on the interface used. However, the SCI group only reported moderately significant more difficulty when using the sip-puff interface compared to the head array switch. Furthermore, there is a statistically significant difference between perceived task difficulty while using the 3-axis joystick between the spinal cord injured and uninjured groups (Fig.~\ref{fig:difficulty}). The implications of this finding for the adaptable autonomy algorithm is two-fold: (1) this gives credence to the suggestion that the adaptable autonomy paradigm should be designed specifically for the data collected from the motor-impaired group, and (2) it is possible to perform some of the customization in advance based on the type of interface being used.  
\begin{table}[H]
\caption{Factor Correlations with Task Difficulty}
\label{table:corr}
\centering
\begin{threeparttable}
  \begin{tabular}{|c|c|c|c|c|}
    \cline{2-5}
    \multicolumn{1}{c}{}                 &\multicolumn{4}{|c|}{\textbf{Task Difficulty}}                        														\\ \cline{2-5}
    \multicolumn{1}{c}{}                 &\multicolumn{2}{|c|}{\textbf{SCI}}                        &\multicolumn{2}{c|}{\textbf{Uninjured}}                 \\ \cline{2-5}
    \multicolumn{1}{c|}{}                & R                      & p-value                           & R                  & p-value                         \\ \hline
    \textbf{HRV}                         & -0.1429                  & 0.1766                        & 0.0963              & 0.2456                           \\ \hline
    \textbf{User Input Frequency}        & \textbf{0.4155}          & \textbf{\textless0.0001}      & \textbf{0.4516}     & \textbf{\textless0.0001}         \\ \hline
    \textbf{\# Mode Switches}            & 0.2562                   & 0.0142                        & \textbf{0.4940}     & \textbf{\textless0.0001}         \\ \hline
    \textbf{Input Smoothness}            & \textbf{-0.3696}         & \textbf{\textless0.001}       & \textbf{-0.4760}    & \textbf{\textless0.0001}         \\ \hline
    \textbf{Trajectory Smoothness}       & \textbf{-0.4725}         & \textbf{\textless0.0001}      & \textbf{-0.6483}    & \textbf{\textless0.0001}         \\ \hline
    \textbf{Task Completion Time}        & 0.2886                   & 0.005                         & \textbf{0.5521}     & \textbf{\textless0.0001}         \\ \hline
    \end{tabular}
\end{threeparttable}
\end{table}

\subsection{Physiological Measures}\label{sec:results_physiological} 
\begin{wrapfigure}{r}{0.35\textwidth}
\begin{center}
\includegraphics[width=0.3\textwidth]{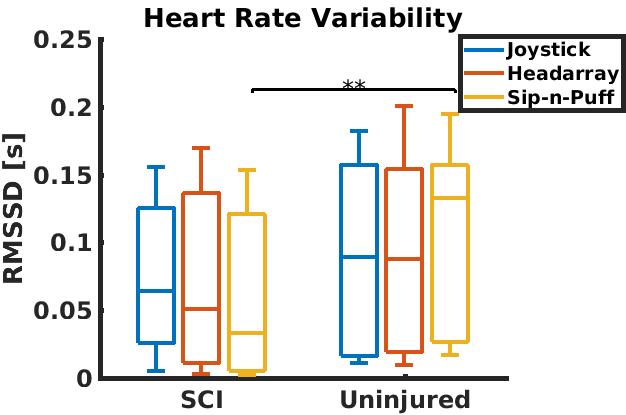}
\caption{Heart rate variability data averaged over subjects and tasks shows significant differences between subject groups for Sip/Puff interface.}
\label{fig:hrv}
\end{center}
\end{wrapfigure}

Our analysis of the heart rate variability data showed no evidence of any correlations with task difficulty (Tbl~\ref{table:corr}, row HRV).

As seen in Figure~\ref{fig:hrv} however, there are significant differences in heart rate variability between the two subject groups while using the Sip/Puff interface. The implications of this for the adaptable autonomy algorithms are: (1) physiological measure of heart-rate variability may not be as informative in shedding light on other metrics which will be measurable in real world operation, and
(2) additional support for the first implication from Section~\ref{sec:results_subjective} that the adaptable autonomy paradigm for the target SCI group should be designed specifically from data collected from SCI subjects.

\subsection{User Interaction Measures}\label{sec:results_user}
\begin{figure}[H]
    \centering
    \vspace{0.2cm}
 		\includegraphics[width=0.3\textwidth]{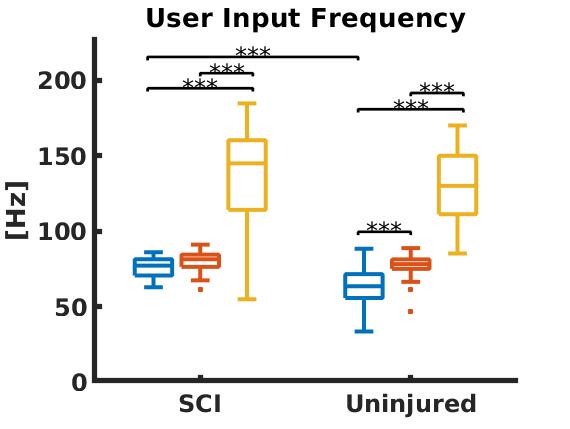}
        \includegraphics[width=0.3\textwidth]{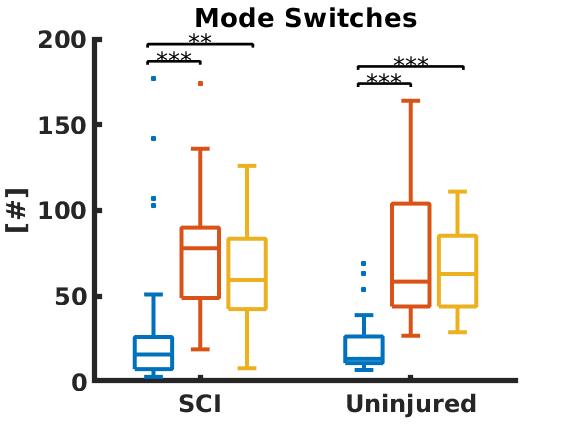}
        \includegraphics[width=0.32\textwidth]{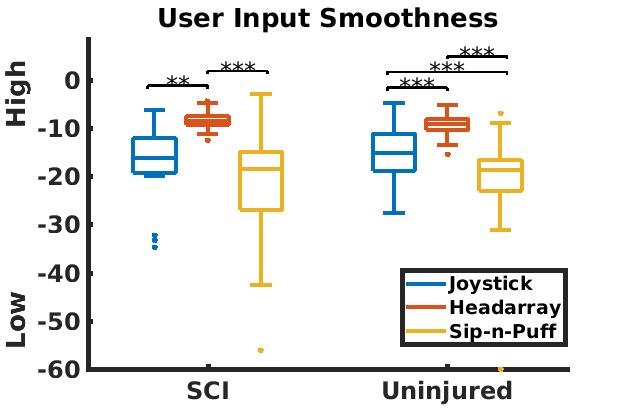}
     \caption{Analysis of user interaction with the robotic arm. \textit{Top:} Frequency of user input commands; \textit{Middle:} Number of mode switches; \textit{Bottom:} Smoothness (continuity) of user input commands.}
\label{fig:user_interact}
\end{figure}

Figure~\ref{fig:user_interact} shows that the user interaction metrics (input frequency, mode switches, input smoothness) are strongly dependent on the type of interface being used. Somewhat surprisingly, with the exception of frequency of input commands using the joystick interface, having a spinal cord injury does not seem to affect how the user provides control commands to the robotic arm. As seen in Table~\ref{table:corr}, the user interaction metrics (rows User Input Frequency, $\#$ Mode Switches, Input Smoothness) can be used to predict task difficulty in both groups---although stronger correlations are observed in the uninjured group. 

The results from analyzing the user interactions metrics have the following implications for the autonomy allocation paradigm: (1) an increase in user input frequency, an increase in the number of mode switches, or a decrease in user input smoothness all can signal an increase in task difficulty---which can be used as a cue for the autonomy to either increase or modify assistance, (2) the second implication of Section~\ref{sec:results_subjective} is corroborated as the results further show it is possible to perform some of the customization in advance by taking into account the control interface in use, and (3) if the autonomy allocation paradigm relies solely on the user interaction metrics, it may be generally acceptable to use a mixture of data from both SCI and uninjured groups to train the model on a larger number of data. 

\subsection{Robot Execution Measure}\label{sec:results_robot}
Table~\ref{table:corr} shows that the smoothness of the end-effector's trajectory is strongly correlated with perceived task difficulty in both groups. As shown in Table~\ref{table:perf},  trajectory smoothness is strongly correlated with task completion times and weakly correlated with successful task completion. This shows that the trajectory smoothness can be an important factor in measuring user performance.

\begin{table}[H]
\centering
\caption{Performance Correlations with Trajectory Smoothness}
\label{table:perf}
\begin{threeparttable}
\begin{tabular}{|l|c|c|c|c|}
\cline{2-5}
\multicolumn{1}{c}{}                 &\multicolumn{4}{|c|}{\textbf{Trajectory Smoothness}}     							       						   \\ \cline{2-5}
\multicolumn{1}{c}{}                 &\multicolumn{2}{|c|}{\textbf{SCI}}                        &\multicolumn{2}{c|}{\textbf{Uninjured}}               \\ \cline{2-5}
\multicolumn{1}{c|}{}                & R                       	& p-value                       & R                   	& p-value                      \\ \hline
\textbf{Success}                     & 0.29                     & \textless0.001                & \textbf{0.36}       	& \textbf{\textless0.0001}     \\ \hline
\textbf{Time}                        & \textbf{-0.71}           & \textbf{\textless0.0001}      & \textbf{-0.81}     	& \textbf{\textless0.0001}     \\ \hline
\end{tabular}
\end{threeparttable}
\end{table}

\begin{figure}[h]
\vspace{0.2cm}
\includegraphics[width=0.4\textwidth]{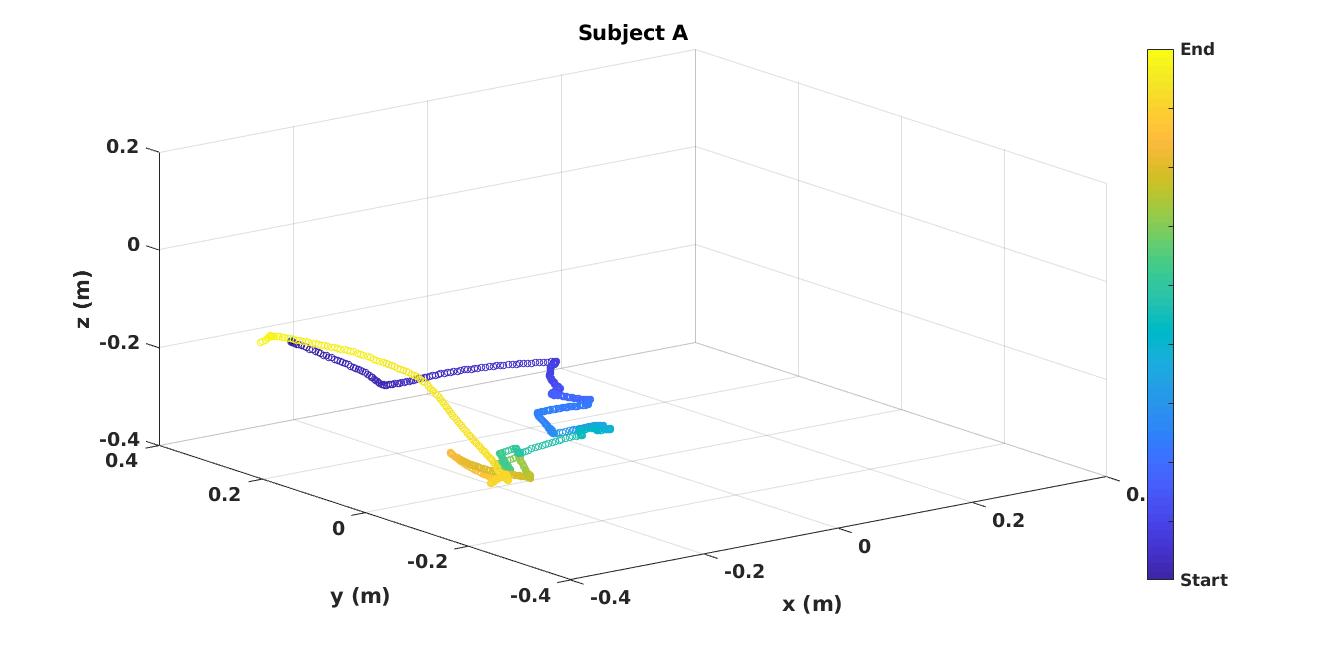}
\includegraphics[width=0.4\textwidth]{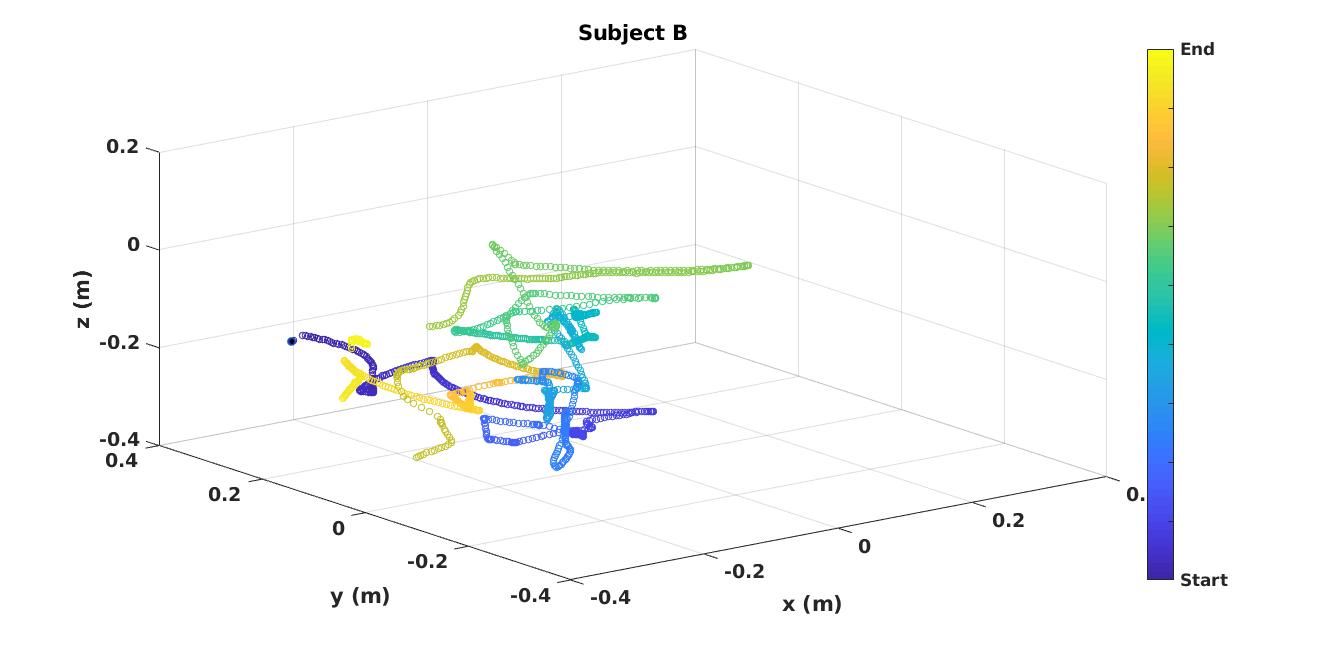}
\centering
\caption{Sample end-effector trajectories from two subjects performing the same task using the same interface.}
\label{fig:traj}
\end{figure}

\begin{wrapfigure}{r}{0.36\textwidth}
\includegraphics[width=0.32\textwidth]{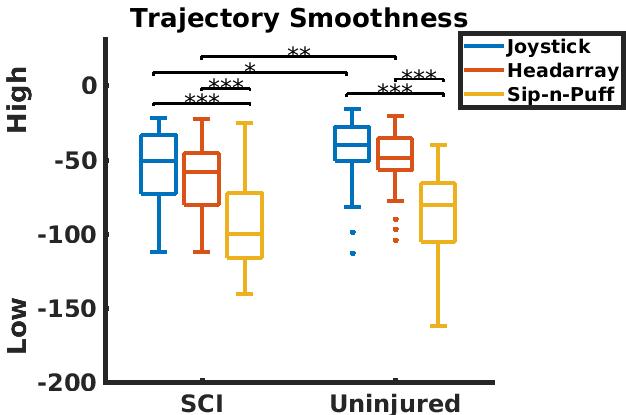}
\centering
\caption{Analysis of end-effector trajectory smoothness between all subjects and interfaces}
\label{fig:robot_exec}
\end{wrapfigure}

We furthermore found great variations in trajectory smoothness within subject groups, interfaces and individuals. For example, Figure~\ref{fig:traj} shows trajectories from two users performing the same task using the same interface with a clear difference in the two trajectories. Figure~\ref{fig:robot_exec} also demonstrates that trajectory smoothness is strongly dependent on the interface used and whether the user has spinal cord injury. 

The implications from the results of analyzing the robot execution metric are: (1) trajectory smoothness can be used as a cue for increasing assistance when smoothness decreases, because not only is the performance quality decreasing, but the user also is experiencing an increase in difficulty, and (2) the second implication from Sections~\ref{sec:results_subjective} and~\ref{sec:results_user} gains further support from these results. 

\section{DISCUSSION}\label{sec:discussion}
The preliminary results can be interpreted as indicative of a relationship between the subjective measure of perceived task difficulty and objective measures that will be available to our assistive autonomy system at runtime during real-world operation---specifically, objective measures that are based on the robot execution (trajectory smoothness) and user interaction with the robot control system (frequency of input commands, number of mode switches, and user input smoothness). Although the preliminary sample size was small, the correlations were strong in the uninjured group, and moderately strong in the spinal cord injured group. These results are encouraging because, unlike perceived task difficulty, the performance measures of user-interaction and robot execution can be computed in real-time and without querying the user. These behavioral and operational metrics may thus be used online to predict whether a user is experiencing changes in the level of perceived difficulty. Additionally, the results showed strong correlations between the robot execution measure of trajectory smoothness and the performance measure of time. These results suggest that each of these factors---frequency of input commands, number of mode switches, user input smoothness, and trajectory smoothness---are potentially useful for modulating autonomy, because they are indicative of user-perceived difficulty.

Heart rate variability is highly age and gender dependent~\cite{Luque-Casado2013}. In general, the heart rate variability of men decreases with age, whereas the heart rate variability of women remains relatively unchanged~\cite{Luque-Casado2016}. In the samples we collected in this preliminary study, there was less variance in age among the uninjured group, and the gender was balanced. In comparison, the uninjured group had a higher mean age with much greater variance, and out of the eight participants there was only one female. The analysis of the heart rate variability data could have been effected by this imbalance. Further experimentation with a larger SCI group size and with gender- and aged-matched controls is necessary. 

One of the aims of this preliminary study was to determine which, if any, of the objective measures were dependent on the injury status of a user. The analysis of the data indicated the measures of task difficulty, frequency of user input commands and robot execution were dependent on whether the user had spinal cord injury. These results imply that an adaptable autonomy paradigm should take into account not only the type of interface, but also should be designed specifically for the target end-user motor-impaired group. This is an important finding as much of the previous research has conducted experiments with mostly expert participants or uninjured participants. Though recruiting spinal-cord injured subjects can be difficult, and the experiments costly and time consuming, data from uninjured subjects will not appropriately inform the design of adaptable autonomy algorithms using the above measures.

An additional aim of the study was to determine how, if at all, the subjective and objective measures changed with the interface used for teleoperation. The data indicated that perceived task difficulty, user interaction measures, and robot execution measures were all dependent on the interface being used to control the robot arm. This implies that some of the customization can occur \textit{a priori} for the control interface in use. 

Our future work will perform more extensive user studies that employ additional  assistive robot platforms and gather data that further characterizes user state according to changes in the user's skill level, task difficulty, and stress levels. Our long term goal is to then use this characterization to inform the design of adaptable autonomy algorithms.

\section{CONCLUSION}\label{sec:conclusion}
In this paper we have presented our preliminary study on characterizing teleoperation commands of people controlling an assistive robotic arm. We investigated the potential usefulness of a variety of performance measures which can be measured in real-time as predictors of whether a user's perceived difficulty in controlling the robotic assistive device is changing. The results of this study found correlations between the objective metrics and the subjectively measured task difficulty, which can inform when the autonomy should step in and assist. 

We also found evidence that the control characteristics of spinal cord-injured participants differed from those of the uninjured participants. This can be interpreted as evidence that to enhance the efficacy of an adaptable algorithm, the training data used must be group specific. We also found that some of the customization can happen \textit{a priori} based on the type of control interface used. 

The next iteration of this work will focus on gathering more data from the target end-user group. Additionally, in our future work we will use the identified measures as cues for when the amount of autonomy in a shared-control assistive system should change. We will evaluate the developed algorithms with end-user subject studies.

\section*{ACKNOWLEDGMENT}

The authors would like to thank Jessica Pedersen for recruiting subjects with spinal cord injury. The authors would also like to thank Chaithanya Krishna Mummidisetty for his valuable comments and helpful suggestions. 

This material is based upon work supported by the National Science Foundation under Grant No. 1552706. Any opinions, findings, and conclusions or recommendations expressed in this material are those of the authors and do not necessarily reflect the views of the National Science Foundation.

\flushbottom

\bibliographystyle{unsrt}  

\end{document}